\documentclass[letterpaper,11pt]{article}

\usepackage{amssymb}
\usepackage{amsmath}
\usepackage{amsthm}
\usepackage{graphicx}
\usepackage{authblk}

\usepackage{float}

\usepackage{booktabs,siunitx}
\usepackage{multirow}

\usepackage{fullpage}
\usepackage{bm}

\usepackage[super]{nth}

\usepackage[utf8x]{inputenc}
\usepackage{color}
\usepackage{algpseudocode, algorithm, algorithmicx}

\usepackage{setspace}

\newcommand\amsclass[1]%
  {\hspace{9mm}
   \textbf{AMS subject classifications. }#1
}

\usepackage[T1]{fontenc}
\usepackage{romannum}
\usepackage{lineno}

\title{Hierarchical Learning to Solve Partial Differential Equations Using Physics-Informed Neural Networks}
\author{Jihun Han\thanks{jihun.han@dartmouth.edu} and Yoonsang Lee\thanks{yoonsang.lee@dartmouth.edu}}
\affil{Department of Mathematics, Dartmouth College}
\date{}

\begin{document}
\pagenumbering{arabic}

\maketitle
\begin{abstract}
The neural network-based approach to solving partial differential equations has attracted considerable attention due to its simplicity and flexibility in representing the solution of the partial differential equation. In training a neural network, the network learns global features corresponding to low-frequency components while high-frequency components are approximated at a much slower rate. For a class of equations in which the solution contains a wide range of scales, the network training process can suffer from slow convergence and low accuracy due to its inability to capture the high-frequency components. In this work, we propose a hierarchical approach to improve the convergence rate and accuracy of the neural network solution to partial differential equations. The proposed method comprises multi-training levels in which a newly introduced neural network is guided to learn the residual of the previous level approximation. By the nature of neural networks' training process, the high-level correction is inclined to capture the high-frequency components. We validate the efficiency and robustness of the proposed hierarchical approach through a suite of linear and nonlinear partial differential equations.
\end{abstract}



\section{Introduction}\label{sec:1.Introduction}

Data-driven methods with neural networks have achieved remarkable success in solving a wide range of partial differential equations arising in various fields \cite{sirignano2018dgm,han2018solving,cai2021physics,berg2019data}. The representation power and trainability of a network are advantageous to represent an arbitrary function, which can be a solution to a PDE. Also, the efficient calculation of the derivatives of a neural network enables one to represent the given law of physics written in a differential operator. In particular, the methods do not require mesh or grid designs, which is useful for high-dimensional problems.

Many research efforts have focused on well-designed objective or loss functions to guide a neural network to approximate the solution of a PDE. An objective function measures how well a neural network satisfies the PDE, typically defined as the empirical mean of the residual by a neural network. Physics-informed neural networks (PINN) \cite{raissi2019physics}, and DGM \cite{sirignano2018dgm} consider the direct PDE residual as the loss function so that the neural network satisfies the PDE in the domain. The approaches in \cite{yu2017deep, karumuri2020simulator} reformulate an elliptic PDE using an equivalent energy minimization problem to train a neural network. For time-dependent problems, the approach proposed in \cite{han2018solving} estimates the solution of parabolic PDEs at a single point by a neural network catered to the time discretization of equivalent backward stochastic differential equations (BSDE). For elliptic PDEs, the method in \cite{han2020derivative} utilizes a stochastic representation in solving elliptic PDEs and trains a neural network in the form of reinforcement learning.

In particular, PINN has flexibility in informing physical laws described in differential equations, and thus it has been employed in solving a wide range of PDEs. Despite its successful results in many applications, PINN suffers from a slow convergence rate and accuracy degradation for a certain class of PDEs. Recent works have enlightened the limitations of a multi-objective optimization problem comprising different loss components corresponding to the governing differential equation and initial/boundary conditions. The work in \cite{wang2021understanding} inspected the gradients of individual loss components during the training process. The result demonstrated the undesirable pathology of stiff gradient flows caused by the imbalance between the different loss components during back-propagation and showed that the imbalance could degrade the overall training process of the neural network.

The training dynamics of the standard PINN model have been analyzed in \cite{wang2020and} using the neural tangent kernel (NTK) theory \cite{jacot2018neural}. The analysis shows that the eigenvalues of NTK can estimate the convergence rate of different loss components. The authors address the discrepancy in convergence rates as a fundamental reason for the degradation of the convergence rate in the overall test error. Adaptive weighting on loss components was proposed in the sense of convergence rate normalization to mitigate the discrepancy \cite{van2020optimally}. The study approached the balance between loss components in the perspective of the relative error. Motivated by the fact that approximation of derivatives tends to be highly correlated to the magnitude of true derivatives, an optimal choice of the weighted loss function is derived under the full knowledge of the true solution followed by a heuristic method called magnitude normalization.

Such computational challenges are often inherent from the characteristics of the solution of a PDE, in particular when the solution involves a wide range of scales. The multiscale PDE problems arise in various scientific domains, such as fluid dynamics, quantum mechanics, or molecular dynamics. Standard methods, such as finite difference methods (FDM) or finite element methods (FEM), encounter an intractable computational complexity in resolving all relevant scales, numerical instabilities, or slow convergence in general. There have been significant efforts in developing efficient discretization methods for multiscale problems. As a representative example, the multigrid (MG) method \cite{briggs2000multigrid} addresses the disparate convergence rates of different scale components through a hierarchical design of discretizations. The MG method captures the diverse target scale components of the solution from the collaboration of scale-corresponding grid approximations. The MG method achieves fast convergence as the method approximates all scale components corresponding to the grids. A hierarchical approach for multiscale problems has also been discussed in \cite{leeengquist} for turbulent diffusion. Instead of using a fine resolution grid for the whole domain at each level, the approach in \cite{leeengquist} uses a local spatiotemporal domain. By designing a hierarchy that captures all possible scale ranges of the solution, the approach can capture the effective macroscopic behavior by a significant computational gain.

Neural network-based methods also face hurdles in approximating the multiscale solution of a PDE in that a neural network prefers low frequencies (F-Principle) \cite{xu2019frequency}. Therefore, a standard network design can be ineffective in learning the high-frequency components. There are several recent research efforts to address the limitation in training high-frequencies by modifying the architecture or the ingredients of neural networks \cite{cai2019multi, tancik2020fourier, wang2021eigenvector}. In particular, \cite{wang2021eigenvector} proposed a neural network structure with Fourier feature embeddings to learn the multiscale solution efficiently. The embedding allows one to specify target characteristic frequencies of the neural network. The authors consider the multiple embeddings of inputs to simultaneously learn the diverse range of frequencies in the solution.

This work proposes hierarchical learning for solving PDEs to expedite the convergence through a hierarchical design of neural networks. We aim to decompose the learning by multiple target segments of the frequency spectrum and combine all learning consequences to the overall approximation of the PDE solutions. Each reduced target could be resolved efficiently. We impose a hierarchy in neural networks, each corresponding to the different target characteristic frequencies. The neural networks are trained in sequence to correct the residual of the approximation up to previous levels. By F-Principle, the targets of neural networks are naturally aligned from low to high-frequency components in the solution.  We employ the Fourier feature embedding to learn each target frequencies efficiently. With the support of other techniques to improve the convergence, such as the aforementioned adaptive weighting algorithms, the proposed hierarchical method accelerates the convergence of the training process. This study investigates the effect of hierarchical learning in the framework of the Physics-informed neural network. We believe that the idea could be applied to the other neural network-based methods, which we leave as future work.

The rest of the paper is organized as follows. Section \ref{sec:2.PINN} reviews the PINN method and discusses the previous efforts to overcome the spectral barriers in training neural networks to solve PDEs. In section \ref{sec:3.HiPINN}, we propose the hierarchical learning methodology to solve PDEs using standard MLPs, and Fourier feature embedded neural networks. Section \ref{sec:4.Results} provides numerical experiments of linear and nonlinear PDEs validating the performance of proposed methods. Finally, we conclude this paper with discussions about the limitation and future directions of the current study in section \ref{sec:5.Discussion}.

\section{Physics-informed neural networks}\label{sec:2.PINN}
In this section, we summarize the standard Physics-informed Neural Networks (PINN) for a boundary value problem and discuss its variants to address the limitations of PINN. We consider the partial differential equation of unknown real-valued function $u$ in a bounded domain $\Omega \subset \mathbb{R}^n$ 
\begin{align}\label{eq:givenPDE}
\begin{split}
\mathcal{N}[u](\bm{x}) &= f(\bm{x}), ~\bm{x} \in \Omega, \\
\mathcal{B}[u](\bm{x}) &= g(\bm{x}), ~\bm{x} \in \partial \Omega,
\end{split}
\end{align}
where $\mathcal{N}$ is a differential operator and $\mathcal{B}$ represents a boundary condition operator, such as Dirichlet, Neumann, periodic boundary conditions, or a mixed form of them.


General deep learning-based methods to solve Eq.~\eqref{eq:givenPDE} employ a neural network, $u(\bm{x};\bm{\theta})$, to approximate the solution, and train the parameters $\bm{\theta}$  under the guidance of a loss function leading the neural network to satisfy Eq.~\eqref{eq:givenPDE}.
The PINN measures the direct PDE residuals in the loss function\begin{equation}
\mathcal{L}(\bm{\theta}) = \lambda_{\Omega}\mathcal{L}_{\Omega}(\bm{\theta}) + \lambda_{\partial \Omega}\mathcal{L}_{\partial \Omega}(\bm{\theta}),
\end{equation}
which consists of the interior and boundary loss terms
\begin{equation}
 \mathcal{L}_{\Omega}(\bm{\theta}) = \frac{1}{N_{r}}\sum \limits_{i=1}^{N_r} \left|\mathcal{N}[u(\cdot;\bm{\theta})](\bm{x}_r^{i}) - f(\bm{x}_r^{i})\right|^2
\end{equation}
and
\begin{equation}
  \mathcal{L}_{\partial \Omega}(\bm{\theta}) =  \frac{1}{N_b}\sum \limits_{i=1}^{N_b} \left|\mathcal{B}[u(\cdot;\bm{\theta})](\bm{x}_b^{i}) - g(\bm{x}_b^{i})\right|^2
\end{equation}
respectively. 
Here $\left\{\bm{x}^{i}_r\right\}_{i=1}^{N_r}$ and $\left\{\bm{x}^{i}_b\right\}_{i=1}^{N_b}$ are sampling points in the interior, and the boundary of the domain, respectively. Both loss terms can be understood as Monte-Carlo approximation of the constraints that constitute the PDE model. We note that the derivatives of the neural network with respect to the input $\bm{x}$ and the parameters $\bm{\theta}$ can be efficiently computed by the automatic differentiation (AD) \cite{baydin2018automatic}, which is the strength in using neural networks in general. The hyperparameters $\lambda_{\Omega}$ and $\lambda_{\partial \Omega}$ determine the balance between the two-loss terms, which can be determined by the ideas in \cite{wang2021understanding,wang2020and,van2020optimally}.
The neural network is trained by minimizing the loss function with a gradient descent method as 
\begin{equation}
\bm{\theta}_n = \bm{\theta}_{n-1} - \alpha \nabla_{\bm{\theta}} \mathcal{L}(\bm{\theta}).
\end{equation}
The learning rate $\alpha$ could be tuned on each step. The gradient step can also be modified by reflecting the previous steps such as Adam optimization \cite{kingma2014adam}. Moreover,  the sampling points in the interior and on the boundary can be chosen randomly at every iteration when considering the stochastic gradient descent method (SGD).

Despite the remarkable achievement in many applications, the PINN often struggles to learn the solutions of PDEs with either slow convergence or degraded accuracy. Recent works have endeavored to understand unfavorable training scenarios of neural networks and proposed alternative methodologies to overcome the limitations. One of the methods includes balancing different terms of the loss function in the context of multi-objective optimization discussed in section \ref{sec:1.Introduction} \cite{wang2021understanding,wang2020and,van2020optimally}. Another direction addresses the intrinsic behavior of training neural networks,  which is specifically disadvantageous to learn functions involving diverse frequency spectrum \cite{arpit2017closer,rahaman2019spectral,xu2019frequency,xu2019training}. 

The general learning process of neural networks has been studied from spectral analysis  \cite{arpit2017closer,rahaman2019spectral,xu2019frequency,xu2019training}. The F-principle \cite{xu2019frequency} shows that the gradient-based training process has spectral bias as the neural networks tend to learn low frequencies while it requires a longer time to fit high frequencies.  This phenomenon is a challenge in neural network-based methods to solve multiscale PDE problems that suffer from slow convergence or low accuracy. The networks miss the high-frequency components unless the training process is sufficiently long to learn high-frequency components.
\cite{cai2019multi} proposed a neural network architecture with an input scaling treatment for converting the high-frequency components to low-frequency ones preferable to learning. The network is installed with a compact supported activation function and is effectively applied to multiscale applications \cite{wang2020multi, liu2020multi}.
\cite{jagtap2020adaptive} introduced adaptive activation functions with trainable scaling factors in considering the dynamical topology of the loss function in an optimization process.
The authors empirically demonstrate the effect of the adaptive activation function in the frequency domain to accelerate the convergence and improve the accuracy.

Another work in \cite{tancik2020fourier} showed that a simple random Fourier feature embedding of inputs enables a standard MLP to learn high-frequency components more efficiently in applications of computer vision and graphics. Namely, the embedding corresponds to a map from the input $\bm{x}\in \mathbb{R}^{n}$ to the $2m$-dimensional frequency domain as 
\begin{equation}\label{eq:fourier embedding matrix}
\bm{x} \in \mathbb{R}^{n} 
~~\mapsto~~ 
\left[
\begin{matrix}
\bm{a}\odot\cos(\bm{B}_{\sigma}\bm{x}) \\
\bm{a}\odot\sin(\bm{B}_{\sigma}\bm{x}) 
\end{matrix}
\right] \in \mathbb{R}^{2m},
\end{equation}
where $\bm{B}_{\sigma}\in \mathbb{R}^{n \times m}$ is a random wave number matrix sampled from the Gaussian distribution $\mathcal{N}(0, \sigma^2)$ and $\bm{a} \in \mathbb{R}^{m}$ is a scaling vector. The authors analyzed the effect of the embedding on the neural tangent kernel (NTK) of the standard MLP to attenuate the spectral bias using appropriate $\sigma$ and $\bm{a}$.
The Fourier feature embedding is extended to solve multiscale PDE problems with PINN models in \cite{wang2021eigenvector}. The authors proposed the multiple Fourier feature embeddings 
of inputs rather than a single embedding to learn diverse frequency components simultaneously. Moreover, it was also demonstrated that separation in spatial and temporal embeddings effectively handles the problems with different multiscale behavior in spatial and temporal directions.


\section{Hierarchical PINN}\label{sec:3.HiPINN}
The methods discussed in the previous section focus on various strategies to improve the capability of a single neural network to learn a wide range of scales in the solution of a PDE. In the current study, we propose a hierarchical design of neural networks to represent the multiscale solution. The proposed method, which we call `hierarchical Physics-informed neural network' (HiPINN), uses a sequence of neural networks to represent different scale components of the PDE solution. Our hierarchical approach is motivated by the multigrid method that uses a hierarchy of different grid sizes to expedite the convergence of an iterative method to solve PDEs \cite{briggs2000multigrid}. The rationale of the multigrid method is that a grid size has its characteristic scale with its corresponding convergence rate. The multigrid method achieves a fast convergence rate by capturing different scale components through variable grid sizes. The idea of the proposed hierarchical approach for PINN is to impose a hierarchy in networks so that each network can capture its corresponding scales, and we uniformly capture all possible ranges of scales.

\subsection{Hierarchical design of networks}
\begin{figure}[t]
\centering
\includegraphics[width=0.8\textwidth]{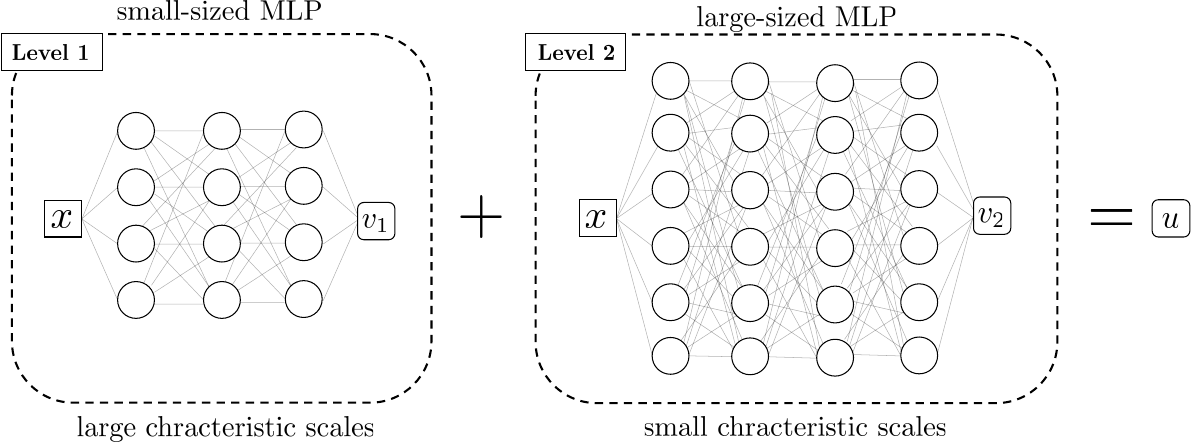}
\caption{Hierarchical composition of standard MLPs; small-sized MLP for capturing low variability components and large-sized MLP for high variability components.}
\label{fig:HiPINN_diagram_MLP}
\end{figure}

HiPINN employs a set of $M$ neural networks $\{v_m(\bm{x};\bm{\theta}_{m})\}_{m=1}^{M}$ with a hierarchy to represent the PDE solution where $M$ represents the number of levels for different characteristic scales. To mimic the hierarchy of the multigrid method, we consider two approaches in the current study. The first approach is the standard multiplayer perceptron (MLP) with various network sizes and complexity. We expect that a simple network will be enough to approximate for low variability components of the unknown solution, while high variability components require a more complicated network. Following this intuitive argument, we increase the complexity of networks by increasing the depth and width of each network. One issue of this approach is that it is unclear to cover a specific range of scales. Suppose two networks are significantly different in terms of complexity. In that case, we expect that the two networks will represent different scale components, but it is not clear whether there is a gap between them. In the study of the spectral bias of neural networks \cite{rahaman2019spectral}, it is shown that higher frequencies are significantly less robust than lower ones in the perturbation of the neural network parameters. This observation indicates that a limited volume in the parameter space is involved in expressing the high-frequency components. With the support of this observation, we consider the complexity of networks in composing the hierarchy. The schematic diagram of the hierarchical employment of the MLPs is displayed in Fig.~\ref{fig:HiPINN_diagram_MLP}.

\begin{figure}[t]
\centering
\includegraphics[width=1.0\textwidth]{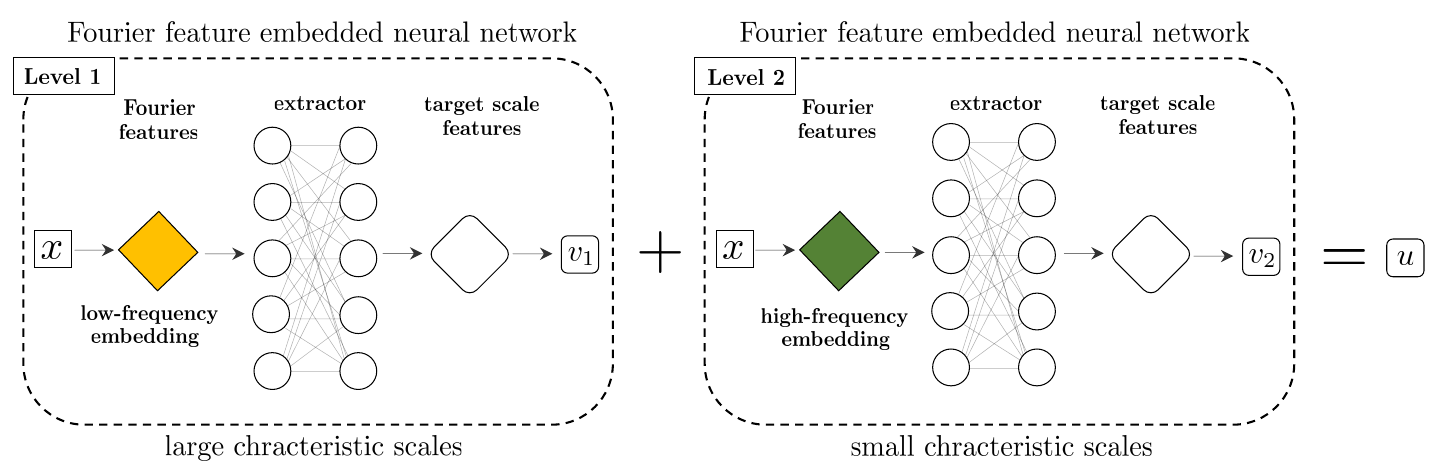}
\caption{Hierarchical composition of Fourier feature embedded neural networks; With the same network architecture, the target characteristic frequency is controlled by the Fourier feature embedding of inputs as Eq.~\eqref{eq:fourier embedding matrix}.}
\label{fig:HiPINN_diagram_FFENN}
\end{figure}

Another approach to impose a hierarchy in the network is the Fourier feature embedding \cite{tancik2020fourier}. The structure of each network is identical through levels with the input embedding as Eq.~\eqref{eq:fourier embedding matrix}. However, we vary them by increasing $\sigma$ as the level ($m$) increases so that a high-level network represents high frequency or wavenumber behaviors compared to the ones captured by the low-level networks. As the network size does not change through the hierarchy, the Fourier embedding-based approach does not provide any computational efficiency in solving a low-level network compared to the hierarchy using the network complexity of MLP. However, the Fourier embedded hierarchy can specify the target characteristic scales through $\sigma$.
In the multiscale approach using the Fourier embedding for PINN \cite{wang2021eigenvector}, a various range of $\sigma$ values is incorporated to design a single network to target all possible ranges of scales in the solution. In terms of the network complexity, HiPINN does not necessarily use a network more complicated than the one used in \cite{wang2021eigenvector}.  The goal of HiPINN is to expedite the training process by dividing the training into specific scales instead of training all possible scales simultaneously. The schematic of a hierarchical neural network design using the Fourier feature embedding is shown in Fig.~\ref{fig:HiPINN_diagram_FFENN}.

\subsection{HiPINN algorithm}
 Using the neural networks with a hierarchy, the $M$-level HiPINN representation of the PDE solution is the sum of all neural networks, which is given as 
\begin{equation}
u_M(\bm{x}) = \sum \limits_{m=1}^{M}v_m(\bm{x}; \bm{\theta}_m).
\end{equation}
Under this structure, the training of each level network is on the correction of the residual of the previous level solution representation. To add the $(M+1)$-th level to $u_M$ using $v_{M+1}$, the loss function $\mathcal{L}^{(M+1)}$ is 
\begin{equation}\label{eq:}
\mathcal{L}^{(M+1)}(\bm{\theta}_{M}) = \lambda_{\Omega} \mathcal{L}^{(M+1)}_{\Omega}\left(\bm{\theta}_{M+1}\right) +  \lambda_{\partial \Omega} \mathcal{L}^{(M+1)}_{\partial \Omega}\left(\bm{\theta}_{M+1}\right) ,
\end{equation}
where 
\begin{align}
\begin{split}\label{eq:}
\mathcal{L}^{(M+1)}_{\Omega}(\bm{\theta}_{M+1})&= \frac{1}{N_{r}}\sum \limits_{i=1}^{N_r} \left|\mathcal{N}[u_{M}+v_{M+1}(\cdot;\bm{\theta}_{M+1})](\bm{x}_r^{i}) - f(\bm{x}_r^{i})\right|^2, \\
 \mathcal{L}^{(M+1)}_{\partial \Omega}(\bm{\theta}_{M+1}) &= \frac{1}{N_b}\sum \limits_{i=1}^{N_b} \left|\mathcal{B}[u_{M}+v_{M+1}(\cdot;\bm{\theta}_{M+1})](\bm{x}_b^{i}) - g(\bm{x}_b^{i})\right|^2.
\end{split}
\end{align}
We note that $u_M$ is already approximated, and thus the training variable related to $\mathcal{L}^{(M+1)}$ is $\bm{\theta}_{M+1}$.

If the differential operator and the boundary operator are linear, the $(M+1)$-th level training is equivalent to solving the original PDE operator using $v_{M+1}$ for modified $f^{(M+1)}(\bm{x})$ and $g^{(M+1)}(\bm{x})$, which are given by
\begin{equation}f^{(M+1)}(\bm{x}) = f(\bm{x}) - \mathcal{N}[u_{M}](\bm{x})\end{equation}
and
\begin{equation}g^{(M+1)}(\bm{x}) = g(\bm{x}) - \mathcal{B}[u_{M}](\bm{x}),\end{equation}
respectively. Therefore, the implementation for the linear case involves only marginal modification of the standard PINN method. When the differential operator $\mathcal{N}$ is nonlinear, the differential operator on $v_{M+1}$ at the $(M+1)$-th level will be different from the original operator $\mathcal{N}$. However, the structure of the operator does not change over the level; it remains at minimizing the loss related to $\mathcal{N}[\mbox{`approximation up to the previous level'}+\mbox{`current level network'}]$ over the current level network. Thus HiPINN for a nonlinear problem requires only one implementation of a solver and uses it repeatedly for all levels.

We also note that HiPINN does not require any projection or interpolation operations between different level solutions, which are crucial in the algebraic multigrid method. In HiPINN, each level approximate solution uses the same sampling points $\left\{\bm{x}^{i}_r\right\}_{i=1}^{N_r}$ and $\left\{\bm{x}^{i}_b\right\}_{i=1}^{N_b}$. From the homogeneity of the problem to be solved at each level, it is straightforward to implement various types of cycles to iterate over different levels, such as V and W cycles \cite{briggs2000multigrid}. The V cycle starts from a low resolution to a high resolution and iterates back to a low resolution. The W cycle repeats the V cycle to approximate scale components that are not sufficiently captured at the corresponding level.

\section{Numerical experiments}\label{sec:4.Results}
In this section, we validate the robustness and effectiveness of the proposed hierarchical learning methodology to solve PDEs through a suite of test problems. In all numerical experiments, we use the standard multilayer perceptrons (MLPs) and Fourier feature embedded neural networks \cite{wang2021eigenvector} with the $\tanh$ activation function. In the Fourier feature embedding case, the architecture of each network is designed as follows in sequence; 1) multiple Fourier feature embeddings of input, each of embedding corresponding to the map in Eq.~\eqref{eq:fourier embedding matrix} with scaling vector $\bm{a}=\bm{1}$, 2) a multiscale feature extractor MLP common for each embedded feature, 3) a final linear layer passed by concatenated features extracted. In our numerical experiments, we consider the dimension of a Fourier feature embedding the same as that of the first hidden layer of the multiscale feature extractor. Moreover, we include a dense layer to pass the concatenated features, which performs better than direct linear mapping to the output in our experiments. We train each neural network using the Adam optimizer \cite{kingma2014adam} with the learning parameters $\beta_1=0.95$ and $\beta_2=0.95$, and all the trainable parameters are initialized from Glorot normal distribution \cite{glorot2010understanding}. Moreover, we employ the adaptive weights algorithm \cite{wang2020and} in all experiments, updating the weights in every $100$ gradient descent steps for computational efficiency. Except the first test in which an exact solution is available, we obtain reference solutions using the FEM method with sufficiently large mesh sizes. We measure the accuracy of the network-based solutions $\tilde{u}$ using the relative $\mathcal{L}^2$-error, $\frac{\|\tilde{u}-u\|_{2,\Omega}}{\|u\|_{2,\Omega}}$. All benchmark losses are referred to the test losses computed on the corresponding grid points.

\subsection{Poisson equation}
As the first example, we consider the Poisson equation in the unit square $\Omega=[0,1]^2$ with a Dirichlet boundary condition,
\begin{align}\label{eq:linPoisson}
\begin{split}
\Delta u &= f ~~\textrm{in}~~ \Omega,  \\
u &= g ~~\textrm{on}~~ \partial \Omega.
\end{split}
\end{align}
Here, we choose the force term $f$ and the boundary value $g$ such that Eq.~\eqref{eq:linPoisson} has the exact solution $u(\bm{x})=\sin(8\pi x_1^2+4\pi x_2)\sin(8\pi x_2^2+4\pi x_1)$.
We consider two-level hierarchical learning with neural networks, $v(\bm{x};\bm{\theta}_1)$ and  $v(\bm{x};\bm{\theta}_2)$, which are sequentially trained using the corresponding loss functions,
\begin{align}
(\text{level}~1)~~~~~~~~~~\mathcal{L}^{(1)}(\bm{\theta}_1) &= \frac{\lambda_{\Omega}}{N_r}\sum \limits_{i=1}^{N_r}\left|\Delta v(\bm{x}_r^{i};\bm{\theta}_1) -f(\bm{x}_r^{i}) \right|^2 + \frac{\lambda_{\partial \Omega}}{N_b}\sum \limits_{i=1}^{N_b} \left|v(\bm{x}_b^{i};\bm{\theta}_1) -g(\bm{x}_b^{i}) \right|^2,~~~~~~~ \\
(\text{level}~2)~~~~~~~~~~\mathcal{L}^{(2)}(\bm{\theta}_2) &= \frac{\lambda_{\Omega}}{N_r}\sum \limits_{i=1}^{N_r}\left|\left(\Delta v(\bm{x}_r^{i};\bm{\theta}_1^{\ast})+\Delta v(\bm{x}_r^{i};\bm{\theta}_2)\right) -f(\bm{x}_r^{i}) \right|^2~~~~~~~ \nonumber \\
&~~~ + \frac{\lambda_{\partial \Omega}}{N_b}\sum \limits_{i=1}^{N_b} \left|\left(v(\bm{x}_b^{i};\bm{\theta}_1^{\ast})+v(\bm{x}_b^{i};\bm{\theta}_2)\right) -g(\bm{x}_b^{i}) \right|^2,
\label{eq:level2 loss linPoisson}
\end{align}
respectively. Here, $\bm{\theta}_1^{\ast}$ in Eq.~\eqref{eq:level2 loss linPoisson} is the updated $\bm{\theta}_1$ at level $1$ and is fixed at level 2. We note that as the differential (i.e., Laplacian) and boundary  operators are linear, the second level PDE for the neural network $v(\bm{x};\bm{\theta}_2)$ is also a Poisson equation with a Dirichlet boundary condition with shifted force and boundary functions. We test our method using the standard MLPs, and Fourier feature embedded neural networks with training sample sizes $N_r=400$ and $N_b=400$.

\begin{figure}[!htb]
\centering
\includegraphics[width=1.0\textwidth]{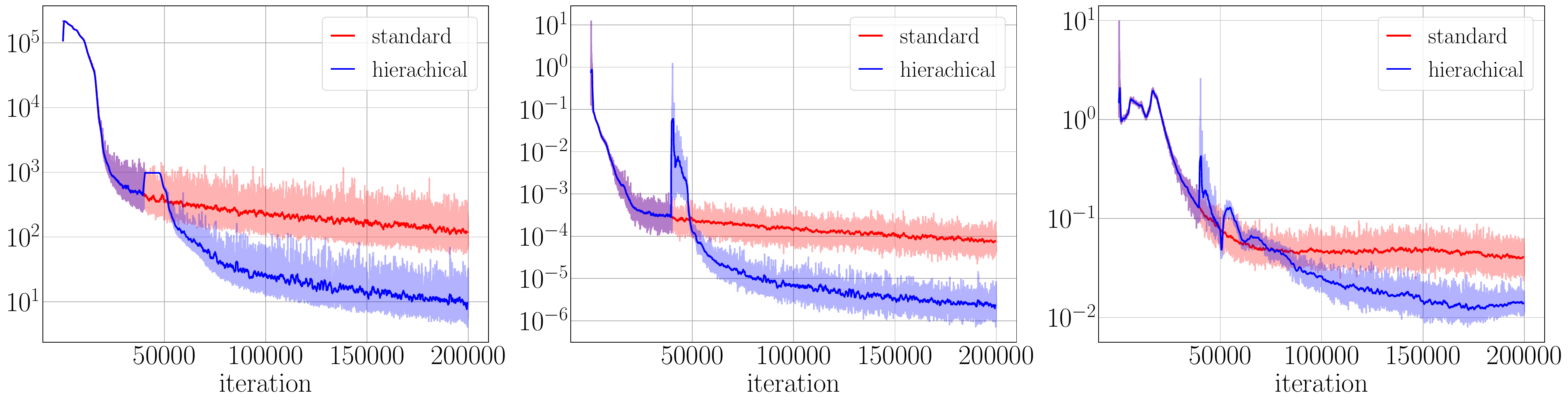}
\caption{Training procedures of  MLPs in solving 2D Poisson equation, Eq.~\eqref{eq:linPoisson}, by standard learning (3 hidden layers,  200 units) and proposed hierarchical learning (first level: 3 hidden layers,  200 units,  second level: 5 hidden layers,  200 units).  (left) interior losses,  (middle) boundary losses, (right) relative $\mathcal{L}^2$-errors.  The test data for each benchmark are obtained from $201\times 201$ uniform grid on $\Omega=[0,1]^2$.  The hierarchical learning corresponds to second level initiation at $4\times 10^4$ iterations.}
\label{fig:linPoisson2D_losses_MLP}
\end{figure}

First, we use the MLP with three hidden layers of dimension $200$ at the first level and  five hidden layers of dimension $200$ at the second level. The hierarchical learning is compared with the standard learning (i.e., single-level hierarchy) with different sizes of MLPs, $H$ numbers of hidden layers of dimension $200$ for $H=2,3,\cdots,8$, among which the MLP with $H=3$ achieves the best performance in approximating the solution. Fig.~\ref{fig:linPoisson2D_losses_MLP} shows the training procedures for $2\times 10^5$ iterations. We observe that the correction at the second level properly works to accelerate the convergence in two losses and achieve better approximation accuracy than the single network learning method.

\begin{figure}[!htb]
\centering
\includegraphics[width=1.0\textwidth]{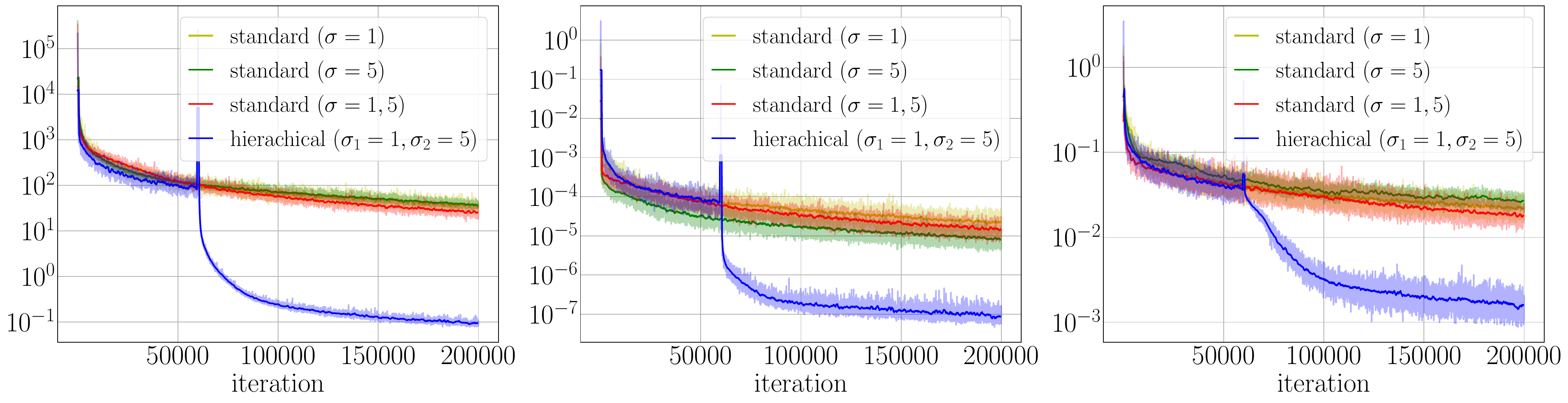}
\caption{Training procedures of Fourier feature embedded neural networks in solving 2D Poisson equation, Eq.~\eqref{eq:linPoisson}, by standard learning and proposed hierarchical learning. (left) interior losses,  (middle) boundary losses,  (right) relative $\mathcal{L}^2$-errors.  The test data for each benchmark are obtained from $201\times 201$ uniform grid on $\Omega=[0,1]^2$. 
The standard learning corresponds to single Fourier feature embedding with $\sigma=1$, $\sigma=5$, separately, and multiple embeddings $\sigma=1,5$. The hierarchical learning runs with single embedding $\sigma=1$ at the first level, and $\sigma=5$ at the second level, in sequence. The second level training is initiated at $6\times 10^4$ iterations. The detail approximations are also presented in Fig.~\ref{fig:linPoisson2D_apprx_FourierEmbedNN}}
\label{fig:linPoisson2D_losses_FourierEmbedNN}
\end{figure}

\begin{figure}[h]
\centering
\includegraphics[width=1.0\textwidth]{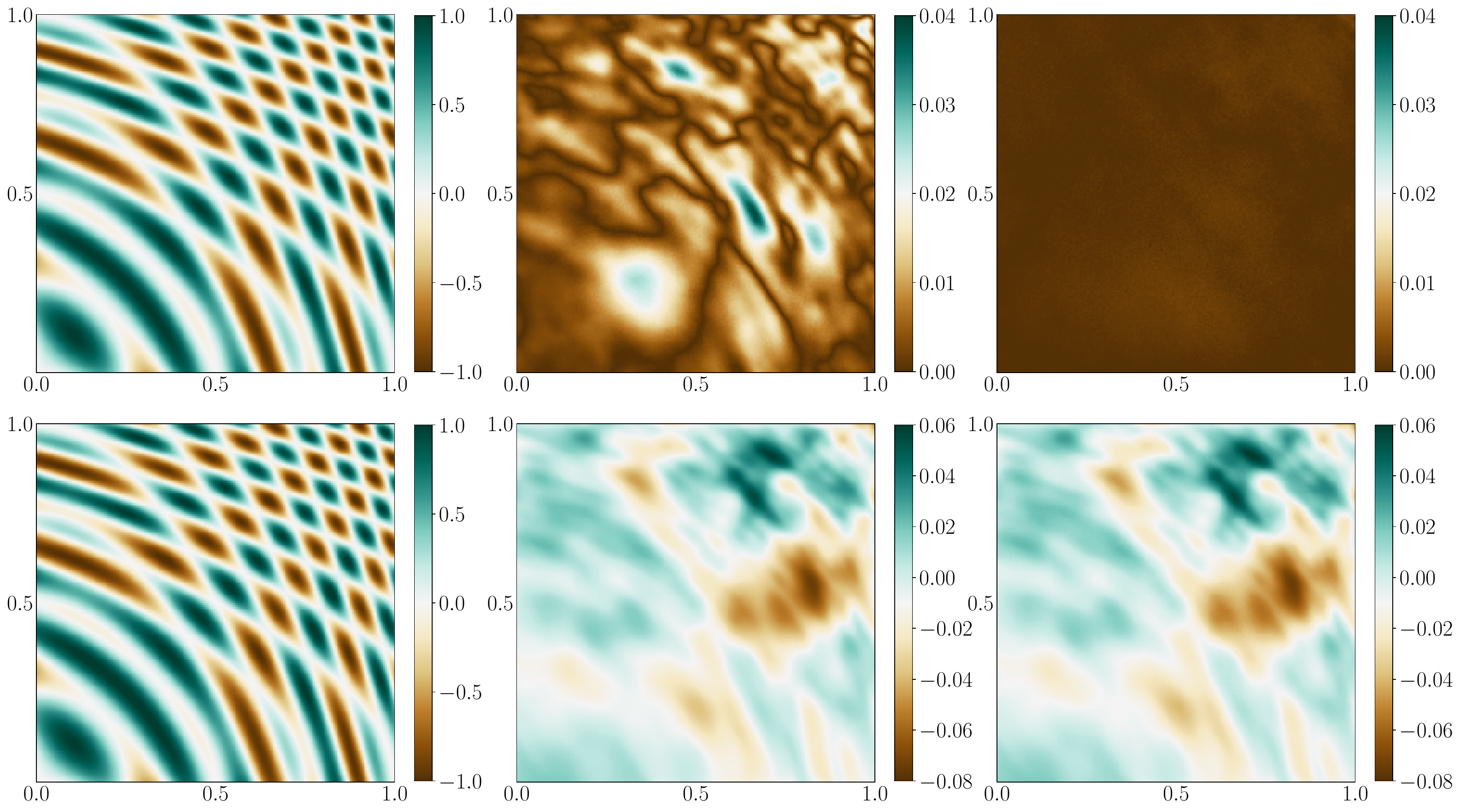}
\caption{first row: The Numerical solutions (Fourier Feature embedded neural networks) of the 2D Poisson equation, Eq.~\eqref{eq:linPoisson} by standard learning (multiple embeddings $\sigma=1,5$) and proposed hierarchical learning (single embedding $\sigma_1=1$, $\sigma_2=5$ in sequence).
(left) the exact solution, (middle) the pointwise error of the approximation from standard learning, (right) the pointwise error of the approximation from proposed hierarchical learning.
second row: the approximations at each level in the hierarchical learning. (left) the approximation $v(\cdot;\bm{\theta}_1^{\ast})$ at the first level, (middle) the approximation $v(\cdot;\bm{\theta}_2^{\ast})$ at the second level, (right) the target function for $v(\cdot;\bm{\theta}_2)$ at the second level, which is equal to ($u_{\text{exact}}-v(\cdot;\bm{\theta}_1^{\ast})$). }
\label{fig:linPoisson2D_apprx_FourierEmbedNN}
\end{figure}

For the case of the Fourier feature embedded neural networks, we use the same size neural networks at both levels, where each network has a different Fourier embedding.  We use single Fourier feature embedding at each level with $\sigma=1$ and $\sigma=5$,  respectively, in considering low target frequencies at the first level and relatively high frequencies at the next level.
The rest of the network consists of the feature extractor with three hidden layers of dimension $200$ followed by the last dense layer of dimension 200. To demonstrate the effectiveness of learning the diverse frequencies from low to high in sequence, we compare our method with the standard learning with the single embedding ($\sigma=1$ and $\sigma=5$) and multiple embeddings ($\sigma=1,5$) aiming to learn various frequencies simultaneously. As shown in Fig.~\ref{fig:linPoisson2D_losses_FourierEmbedNN}, our method accelerates the convergence at the second level and achieves an accurate approximation (relative $\mathcal{L}^2$-error $1.33\times 10^{-3}$) in comparison to the other experiments (best relative $\mathcal{L}^2$-error $1.65\times 10^{-2}$). Fig.~\ref{fig:linPoisson2D_apprx_FourierEmbedNN} shows the point-wise errors of each level approximation in comparison with the standard learning method.
Moreover, our method combined with Fourier feature embedding outperforms the performance of HiPINN using the standard MLP, as we can employ a neural network suitable for learning the target frequencies at each level.

\begin{figure}[b]
\centering
\includegraphics[width=1.0\textwidth]{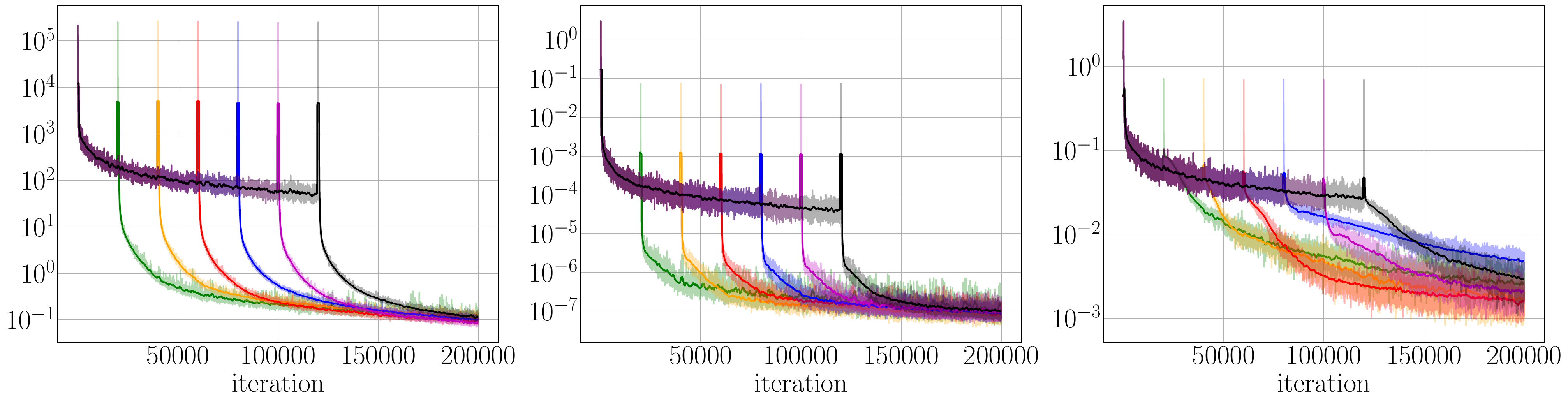}
\caption{Training procedures of Fourier feature embedded neural network (single embedding $\sigma_1=1$, $\sigma_2=5$ in sequence) in solving 2D Poisson equation, Eq.~\eqref{eq:linPoisson} with different second level initiations, $2n \times 10^4$, $n=1,2,3,4,5,6$. (left) interior losses, (middle) boundary losses, (right) relative $\mathcal{L}^2$-errors.}
\label{fig:linPoisson2D_losses_trans_FourierEmbedNN}
\end{figure}

We address a question when it is appropriate to switch to the next level. Fig.~\ref{fig:linPoisson2D_losses_trans_FourierEmbedNN} presents the six training procedures of the Fourier feature embedded neural networks ($\sigma=1$, $\sigma=5$ in level sequence). The experiment shows that transition to the next level after 25000 iterations provide comparable overall accuracy using the two-level representation.

\subsection{Nonlinear Poisson Equation}
For the nonlinear PDEs, the spectral separation could be more complicate than the linear PDEs as the previous level approximation affects the original differential operator. To validate the flexibility of our method along this line, we consider the nonlinear Poisson equation in the unit square $\Omega=[0,1]^2$ with a Dirichlet boundary condition,
\begin{align}\label{eq:nonlinPoisson}
\begin{split}
-\nabla \cdot ( (1+u^2) \nabla u) = f &~~\textrm{in}~~ \Omega, \\
u = g &~~\textrm{on}~~ \partial \Omega.
\end{split}
\end{align}
We choose the force term $f=\frac{1}{2}\exp \left(2+2\sin(10 \pi x_1^2+10\pi x_2) \right)$ to impose a high-frequency behavior while the boundary value is the constant function $g=1$. 

We train neural networks $v(\bm{x};\bm{\theta}_1)$ and $v(\bm{x};\bm{\theta}_2)$ in 2-level hierarchy under the corresponding losses
\begin{align}
(\text{level}~1)~~~~~\mathcal{L}^{(1)}(\bm{\theta}_1) &= \frac{\lambda_{\Omega}}{N_r}\sum \limits_{i=1}^{N_r}\left|\mathcal{R}(\bm{\theta}_1;\bm{x}_r^{i}) \right|^2 + \frac{\lambda_{\partial \Omega}}{N_b}\sum \limits_{i=1}^{N_b} \left|v(\bm{x}_b^{i};\bm{\theta}_1) -g(\bm{x}_b^{i}) \right|^2,~~~~~~~ \\
(\text{level}~2)~~~~~\mathcal{L}^{(2)}(\bm{\theta}_2) &= \frac{\lambda_{\Omega}}{N_r}\sum \limits_{i=1}^{N_r}\left|\mathcal{R}(\bm{\theta}_2;\bm{x}_r^{i}) \right|^2 + \frac{\lambda_{\partial \Omega}}{N_b}\sum \limits_{i=1}^{N_b} \left|\left(v(\bm{x}_b^{i};\bm{\theta}_1^{\ast})+v(\bm{x}_b^{i};\bm{\theta}_2)\right) -g(\bm{x}_b^{i}) \right|^2, \label{eq:level2 loss nonlinPoisson}
\end{align}
where the residuals of PDEs at two levels are 
\begin{align} 
\mathcal{R}(\bm{\theta}_1;\bm{x}_r^{i}) &= -\nabla \cdot ( (1+v(\bm{x}_r^{i};\bm{\theta}_1)^2) \nabla v(\bm{x}_r^{i};\bm{\theta}_1)) - f(\bm{x}_r^{i}), \\
\mathcal{R}(\bm{\theta}_2;\bm{x}_r^{i}) &= -\nabla \cdot ( (1+(v(\bm{x}_r^{i};\bm{\theta}_1^{\ast})+v(\bm{x}_r^{i};\bm{\theta}_2))^2) \nabla (v(\bm{x}_r^{i};\bm{\theta}_1^{\ast}))+v(\bm{x}_r^{i};\bm{\theta}_2))) - f(\bm{x}_r^{i}). \label{eq:level2 PDE residual nonlinPoisson}
\end{align}
Here, $\bm{\theta}_1^{\ast}$ in Eq.~\eqref{eq:level2 loss nonlinPoisson} and Eq.~\eqref{eq:level2 PDE residual nonlinPoisson} is the updated $\bm{\theta}_1$ at level $1$ and is fixed at level 2. In comparison to the linear case, the residual PDE for the neural network $v(\bm{x};\bm{\theta}_2)$ at the second level differs from the original problem with an altered differential operator, in addition to the force and boundary value. As in the previous experiment, we test the effectiveness of our method using standard MLPs, and Fourier feature embedded neural networks with training sample sizes $N_r=900$ and $N_b=300$. 

\begin{figure}[!htb]
\includegraphics[width=1.0\textwidth]{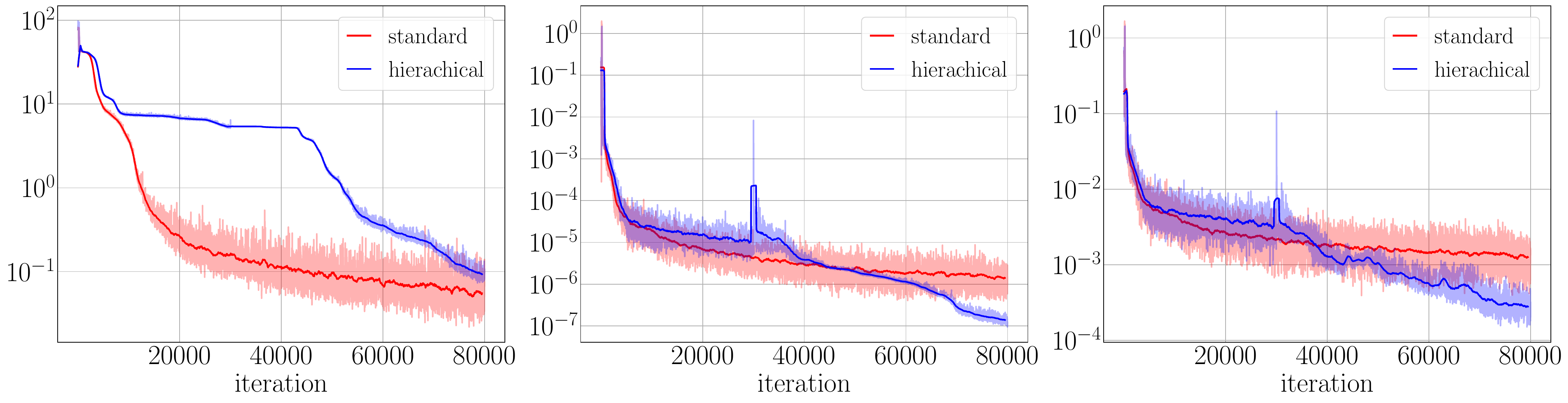}
\caption{Training procedures of  MLPs in solving 2D nonlinear Poisson equation, Eq.~\eqref{eq:nonlinPoisson}, by standard learning (4 hidden layers,  200 units) and proposed hierarchical learning (first level: 3 hidden layers,  200 units,  second level: 5 hidden layers,  200 units).  (left) interior losses,  (middle) boundary losses, (right) relative $\mathcal{L}^2$-errors.  The test data for each benchmark are obtained from $201\times 201$ uniform grid on $\Omega=[0,1]^2$.  The hierarchical learning corresponds to second level initiation at $3\times 10^4$ iterations.
}
\label{fig:nonlinPoisson2D_losses_MLP}
\end{figure}

First, we apply hierarchical learning with MLPs, which comprises three hidden layers of dimension $200$ at the first level and five hidden layers of dimension $200$ at the second level. We train MLPs for $8\times 10^4$ iterations, in which the second level is initiated after $3\times 10^4$ iterations of the first level training. As in the previous test, we compare our method with standard learning using different sizes of MLPs, $H$ numbers of hidden layers of dimension $200$ for $H=2,3,\cdots,8$, which yields the best performance using $H=4$. Fig.~\ref{fig:nonlinPoisson2D_losses_MLP} shows the learning procedure of the two loss terms (interior and boundary losses) and relative $\mathcal{L}^2$-errors. For the $3\times 10^4$ iterations of the first level, the four-layered MLP in standard learning performs better in all benchmarks than the three-layered 
\clearpage
\begin{figure}[!htb]
\includegraphics[width=1.0\textwidth]{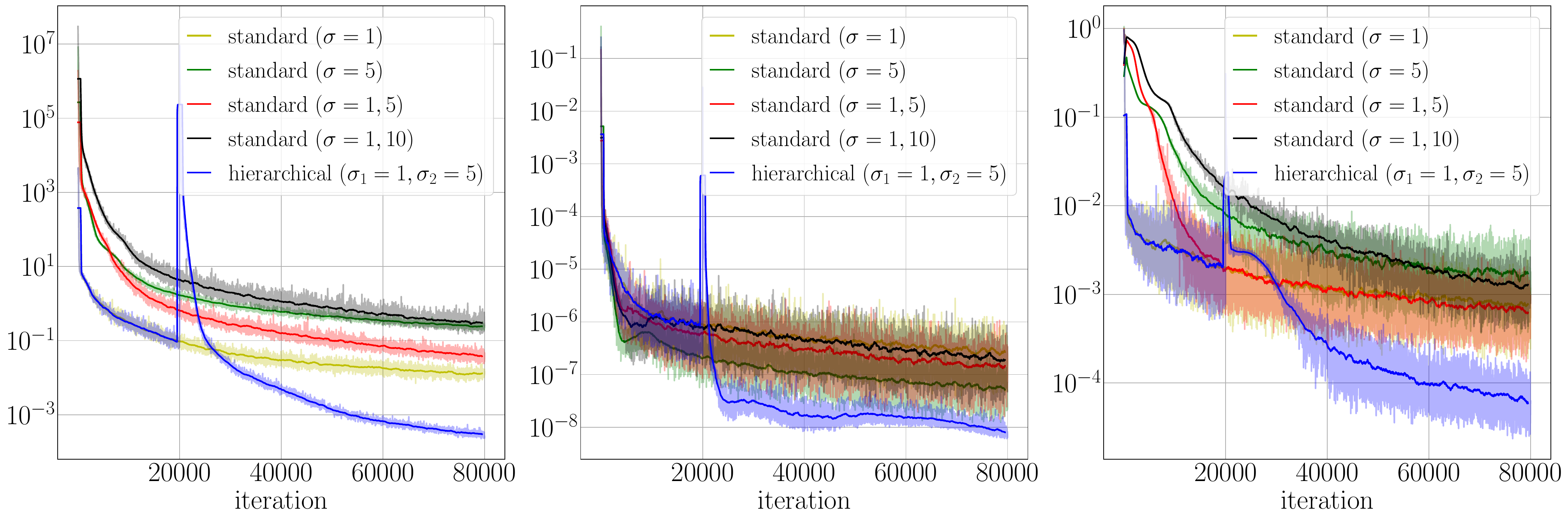}
\caption{Training procedures of Fourier feature embedded neural networks in solving 2D nonlinear Poisson equation, Eq.~\eqref{eq:nonlinPoisson}, by standard learning and proposed hierarchical learning. (left) interior losses,  (middle) boundary losses,  (right) relative $\mathcal{L}^2$-errors.  The test data for each benchmark are obtained from $201\times 201$ uniform grid on $\Omega=[0,1]^2$. 
The standard learning corresponds to single Fourier feature embedding with $\sigma=1$, $\sigma=5$, separately, and multiple embeddings $\sigma=1,5$ and $\sigma=1,10$. The hierarchical learning runs with single embedding $\sigma=1$ at the first level, and $\sigma=5$ at the second level, in sequence. The second level training is initiated at $2\times 10^4$ iterations. The detail approximations are also presented in Fig.~\ref{fig:nonlinPoisson2D_apprx_FourierEmbedNN}
}
\label{fig:nonlinPoisson2D_losses_FourierEmbedNN}
\end{figure}
\noindent MLP in the hierarchical method. Once the five-layered MLP is introduced at the second level, the correction treatment accelerates convergences leading to a more accurate approximation than the four-layered MLP standard learning method.

\begin{figure}[!htb]
\centering
\includegraphics[width=1.0\textwidth]{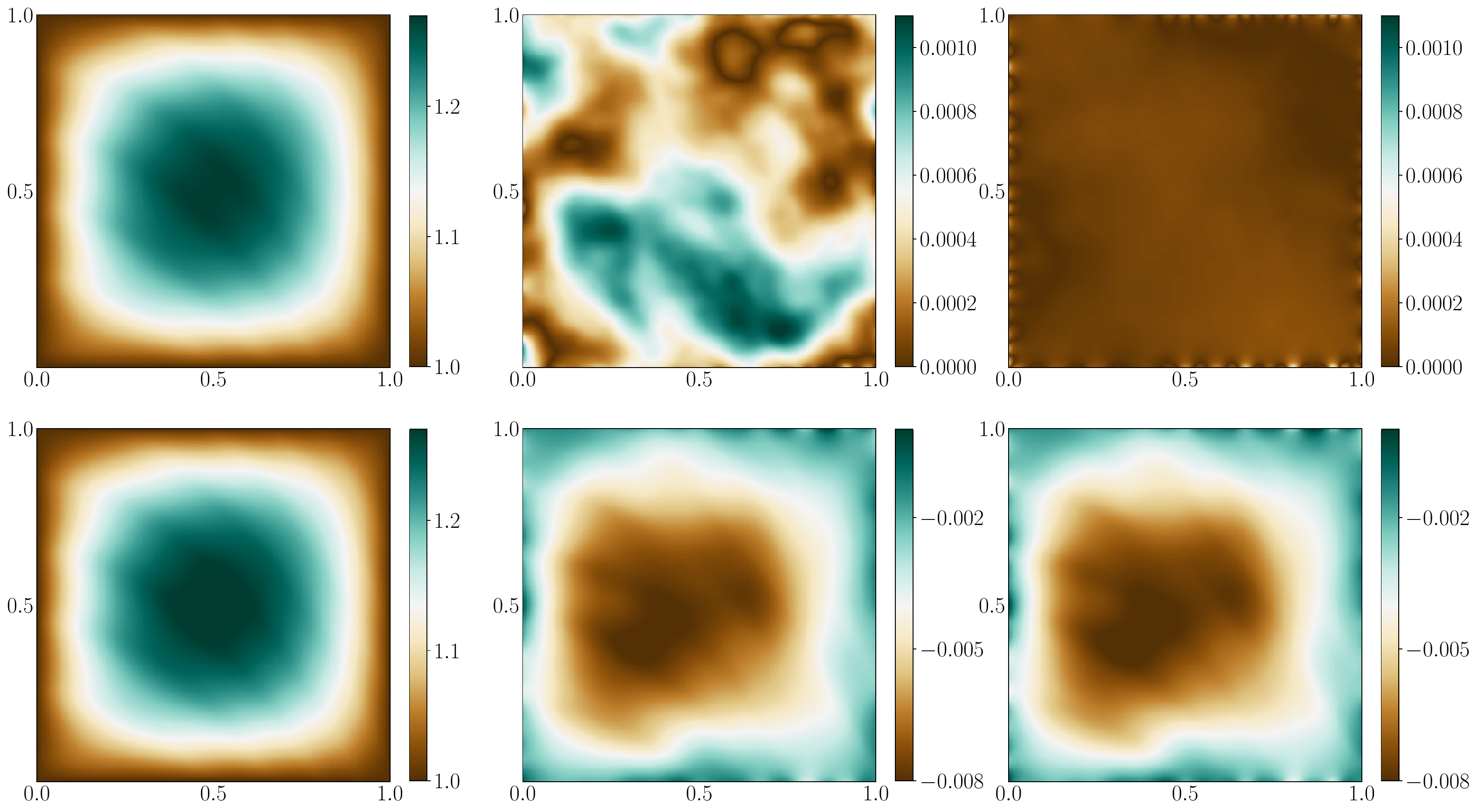}
\caption{first row: The Numerical solutions (Fourier Feature embedded neural networks) of the 2D nonlinear Poisson equation, Eq.~\eqref{eq:nonlinPoisson} by standard learning (multiple embeddings $\sigma=1,5$) and proposed hierarchical learning (single embedding $\sigma_1=1$, $\sigma_2=5$ in sequence).
(left) the exact solution, (middle) the pointwise error of the approximation from standard learning, (right) the pointwise error of the approximation from proposed hierarchical learning.
second row: the approximations at each level in the hierarchical learning. (left) the approximation $v(\cdot;\bm{\theta}_1^{\ast})$ at first level, (middle) the approximation $v(\cdot;\bm{\theta}_2^{\ast})$ at second level, (right) the target function for $v(\cdot;\bm{\theta}_2)$ at second level, which is equal to ($u_{\text{exact}}-v(\cdot;\bm{\theta}_1^{\ast})$). 
}
\label{fig:nonlinPoisson2D_apprx_FourierEmbedNN}
\end{figure}

\begin{figure}[!htb]
\centering
\includegraphics[width=1.0\textwidth]{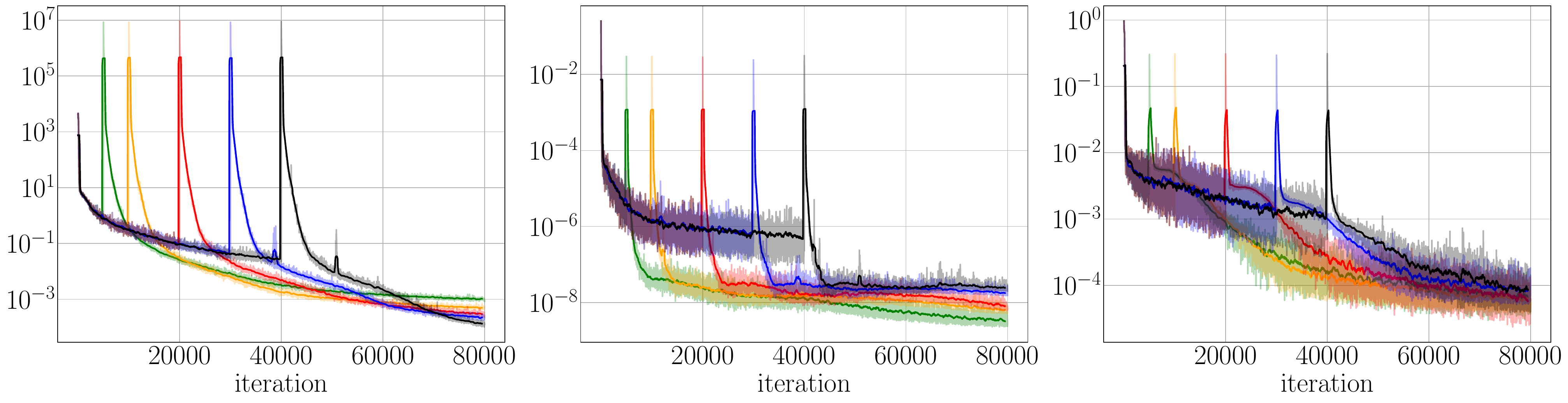}
\caption{Training procedures of Fourier feature embedded neural network (single embedding $\sigma_1=1$, $\sigma_2=5$ in sequence) in solving 2D nonlinear Poisson equation, Eq.~\eqref{eq:nonlinPoisson} with different second level initiations, $n \times 10^4$, $n=0.5,1,2,3,4$. (left) interior losses, (middle) boundary losses, (right) relative $\mathcal{L}^2$-errors.}
\label{fig:nonlinPoisson2D_losses_trans_FourierEmbedNN}
\end{figure}

Next, we employ the neural networks with Fourier feature embeddings to test our hierarchical learning for imposing a hierarchy more explicitly than the standard MLP-based hierarchical learning approach. Using the same size of neural networks at both levels, we consider different single embedding at each level with $\sigma=1$ and $\sigma=5$ respectively, aiming to capture the low-frequency components at the first level and the high-frequency residuals at the second level. The rest of the network is set to the feature extractor with three hidden layers of dimension $200$ followed by the last $200$-dimensional dense layer.  We compare the result to standard learning with different Fourier embeddings; single embedding with $\sigma=1$ and $\sigma=5$, and multiple embeddings with $\sigma=1,5$ and $\sigma=1,10$. We present the training procedures of numerical experiments in Fig.~\ref{fig:nonlinPoisson2D_losses_FourierEmbedNN}. Among the various approaches for  standard learning using a single network, multiple embeddings with $\sigma=1,5$ achieve the best approximation accuracy, $9.72 \times 10^{-4}$ relative $\mathcal{L}^2$-error. On the other hand, the hierarchical learning approach using the Fourier feature embeddings has the best result with a relative $\mathcal{L}^2$-error $6.25 \times 10^{-5}$. (See also Fig.~\ref{fig:nonlinPoisson2D_apprx_FourierEmbedNN}). As we have seen in the linear PDE case, the Fourier feature embedded hierarchical learning shows better performance than standard MLP-based hierarchical learning.
We also conduct the numerical experiments to see the sensitivity of our method to the choice of iteration step to initiate the next level using the Fourier feature embedded neural networks ($\sigma=1$, $\sigma=5$ in level sequence). As shown in Fig,~\ref{fig:nonlinPoisson2D_losses_trans_FourierEmbedNN}, the second level correction works properly to accelerate the convergence and improve the approximation accuracy for all trials with different level iteration settings.

\subsection{Steady-state advection-diffusion equation}
The last test is to demonstrate the capability of the proposed method in handling the multiscale behavior, which could arise from the intertwined consequences of both differential operator and force term. We consider the steady-state advection-diffusion equation with mixed Dirichlet and Neumann boundary conditions,
\begin{align}\label{eq:advDiffusion}
\begin{split}
\bm{w} \cdot \nabla u - \nu \Delta u =  f & ~\textrm{in}~~ \bm{x} \in [0,1]^2, \\
u(0,x_2) = g_1 & ~\textrm{for}~~ x_2 \in [0,1], \\
u(1,x_2) = g_2 & ~\textrm{for}~~ x_2 \in [0,1], \\
\frac{\partial u}{\partial n}u(x_1,0) =\frac{\partial u}{\partial n}u(x_1,1) = 0 &  ~\textrm{for}~~  x_1 \in [0,1].
\end{split}
\end{align}
We choose the diffusion coefficient $\nu=0.01$, the force term $f=\sin(4\pi x_2)$, Dirichlet boundary value, $g_1=0$ and $g_2=1$, and an incompressible velocity field $\bm{w}$,
\begin{align}
\bm{w}(\bm{x}) = 
\left[
\begin{matrix}
-5\sin(6\pi x_1)\cos(6\pi x_2) \\
5\cos(6\pi x_1)\sin(6\pi x_2)
\end{matrix}
\right]. 
\end{align}

We solve Eq.~\eqref{eq:advDiffusion} using two levels, in which neural networks $v(\bm{x};\bm{\theta}_1)$ and $v(\bm{x};\bm{\theta}_2)$ are trained under the following loss functions
\begin{align}
(\text{level}~1)~~~~~\mathcal{L}^{(1)}(\bm{\theta}_1) &= \frac{\lambda_{\Omega}}{N_r}\sum \limits_{i=1}^{N_r}\left|\mathcal{R}(\bm{\theta}_1;\bm{x}_r^{i}) \right|^2 +\frac{\lambda_{\partial \Omega ,1}}{N_{b,1}}\sum \limits_{i=1}^{N_{b,1}} \left|\frac{\partial v}{\partial n}(\bm{x}_{b,1}^{i};\bm{\theta}_1) \right|^2~~~~~~~~~~~~~~~~~~~~~~~~~~~~~~~~\nonumber \\
&~~~+ \frac{\lambda_{\partial \Omega,2}}{N_{b,2}}\sum \limits_{i=1}^{N_{b,2}} \left|v(\bm{x}_{b,2}^{i};\bm{\theta}_1) -g(\bm{x}_{b,2}^{i}) \right|^2,
\end{align}

\clearpage 
\begin{align}
(\text{level}~2)~~~~~\mathcal{L}^{(2)}(\bm{\theta}_2) &= \frac{\lambda_{\Omega}}{N_r}\sum \limits_{i=1}^{N_r}\left|\mathcal{R}(\bm{\theta}_2;\bm{x}_r^{i}) \right|^2+\frac{\lambda_{\partial \Omega ,1}}{N_{b,1}}\sum \limits_{i=1}^{N_{b,1}} \left|\frac{\partial v}{\partial n}(\bm{x}_{b,1}^{i};\bm{\theta}^{\ast}_1) +\frac{\partial v}{\partial n}(\bm{x}_{b,1}^{i};\bm{\theta}_2) \right|^2 ~~~~~~~~~~~~\nonumber \\
&~~~+ \frac{\lambda_{\partial \Omega,2}}{N_{b,2}}\sum \limits_{i=1}^{N_{b,2}} \left|v(\bm{x}_{b,2}^{i};\bm{\theta}^{\ast}_1)+v(\bm{x}_{b,2}^{i};\bm{\theta}_2) -g(\bm{x}_{b,2}^{i}) \right|^2, \label{eq:level2 loss advDiff}
\end{align}
where the residuals of PDE at each level are 
\begin{align} 
\mathcal{R}(\bm{\theta}_1;\bm{x}_r^{i}) &= \bm{w} \cdot \nabla v(\bm{x}_r^{i};\bm{\theta}_1) - \nu \Delta v(\bm{x}_r^{i};\bm{\theta}_1) - f(\bm{x}_r^{i}) \\
\mathcal{R}(\bm{\theta}_2;\bm{x}_r^{i}) &= (\bm{w} \cdot \nabla v(\bm{x}_r^{i};\bm{\theta}^{\ast}_1) - \nu \Delta v(\bm{x}_r^{i};\bm{\theta}^{\ast}_1))+(\bm{w} \cdot \nabla v(\bm{x}_r^{i};\bm{\theta}_2) - \nu \Delta v(\bm{x}_r^{i};\bm{\theta}_2)) - f(\bm{x}_r^{i}). \label{eq:level2 PDE residual advDiff}
\end{align}
Here, $\bm{x}_{b,1}$ and $\bm{x}_{b,2}$ are sampling points for the Neumann and the Dirichlet boundary conditions, respectively, $g$ is read as $g_1$ or $g_2$ depending on the location of $\bm{x}^i_{b,2}$, and $\bm{\theta}_1^{\ast}$ in Eq.~\eqref{eq:level2 loss advDiff} and Eq.~\eqref{eq:level2 PDE residual advDiff} is the updated $\bm{\theta}_1$ at the first level and is fixed at the second level.  Moreover, we also apply the adaptive weight algorithm \cite{wang2020and} by treating the weight $\lambda_{\Omega}$ on boundary loss separately into two parts, $\lambda_{\partial \Omega,1}$ on the Neumann boundary loss and $\lambda_{\partial \Omega,2}$ on the Dirichlet boundary loss.
We note that the second level training $v(\bm{x};\bm{\theta}_2)$ is to approximate the solution of the residual PDE involving the oscillatory advection operator with a modified force term.

We apply the hierarchical learning method with the Fourier feature embedded neural networks. The exact size neural networks are considered at both levels using different single embedding; $\sigma=2$ at the first level and $\sigma=5$ at the second level to learn low and high-frequency components. The rest of the network comprises the feature extractor with three hidden layers of dimension $200$ followed by the last $200$-dimensional dense layer. We train the neural networks over $1\times 10^5$ iterations, where we switch to the second level at various instances (which are at $n \times 10^4$,  $n=1,2,3$,  iterations).

\begin{figure}[!htb]
\centering
\includegraphics[width=1.0\textwidth]{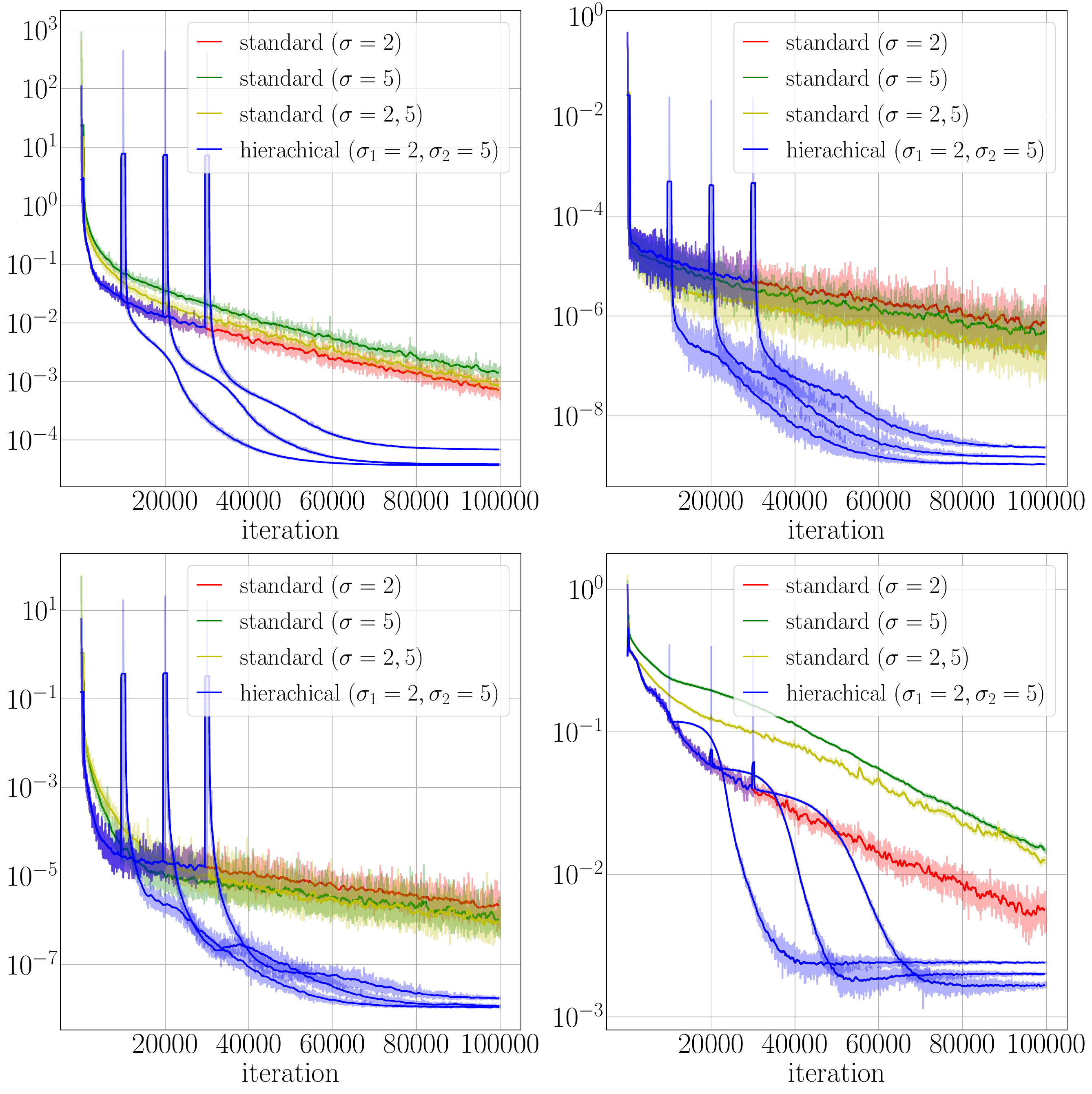}
\caption{Training procedures of Fourier feature embedded neural networks in solving steady-state advection-diffusion equation, Eq.~\eqref{eq:advDiffusion}, by standard learning and proposed hierarchical learning. (first row, left) interior losses,  (first row, right) Dirichlet boundary losses,  (second row, left) Neumann boundary losses,  (second row, right) relative $\mathcal{L}^2$-errors.  The test data for each benchmark are obtained from $201\times 201$ uniform grid on $\Omega=[0,1]^2$. 
The standard learning corresponds to single Fourier feature embedding with $\sigma=2$, $\sigma=5$, separately, and multiple embeddings $\sigma=2,5$. The hierarchical learning runs with single embedding $\sigma=2$ at the first level, and $\sigma=5$ at the second level, in sequence. The second level training is initiated at $n\times 10^4$, $n=1,2,3$, iterations. The detail approximations are also presented in Fig.~\ref{fig:advDiff2D_apprx_FourierEmbedNN}.}
\label{fig:advDiff2D_losses_FourierEmbedNN}
\end{figure}

\begin{figure}[!htb]
\centering
\includegraphics[width=1.0\textwidth]{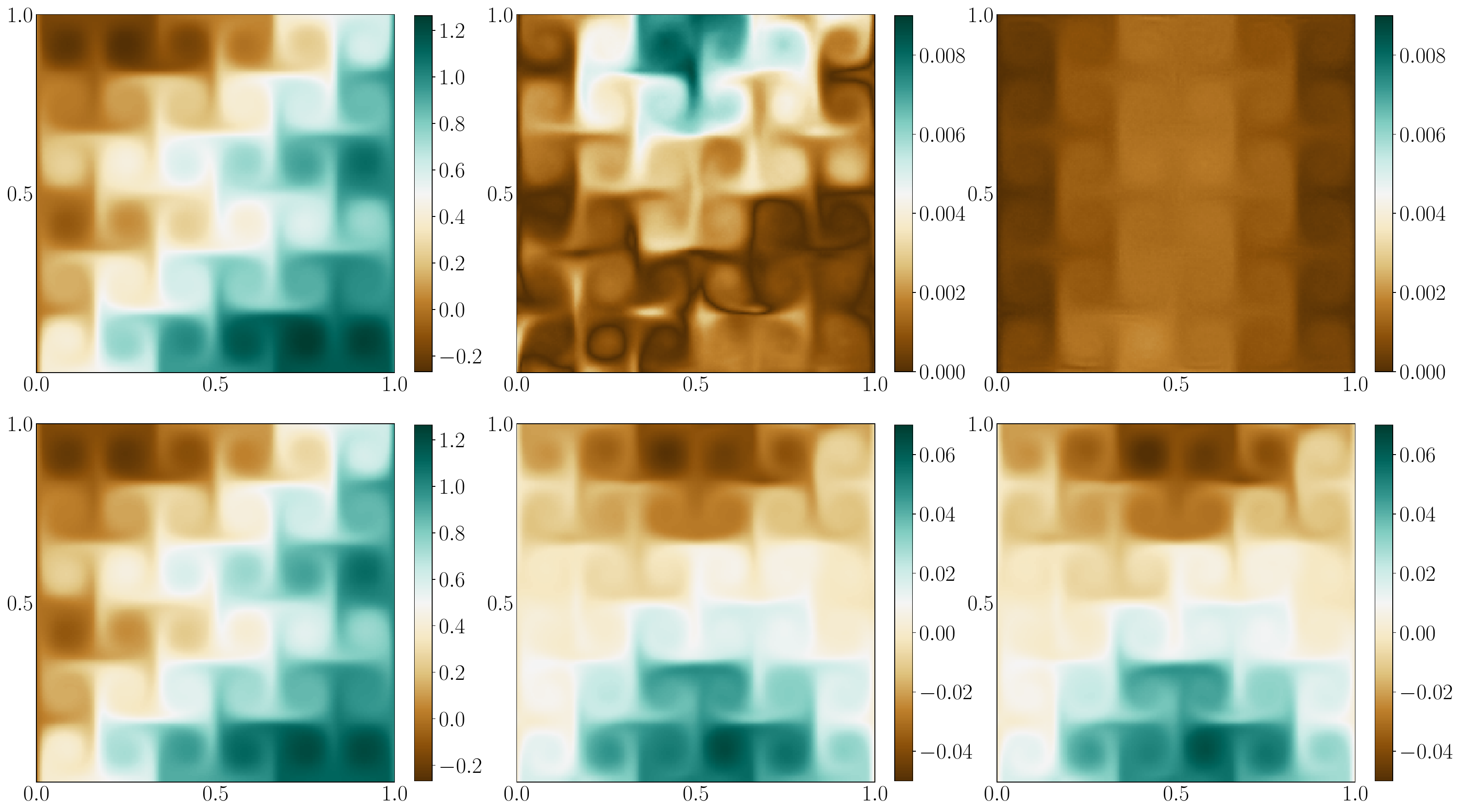}
\caption{
first row: The numerical solutions (Fourier Feature embedded neural networks) of the steady-state advection-diffusion equation, Eq.~\eqref{eq:advDiffusion} by standard learning (single embedding $\sigma=2$) and proposed hierarchical learning (single embedding $\sigma_1=2$, $\sigma_2=5$ in sequence).
(left) the exact solution, (middle) the pointwise error of the approximation from standard learning, (right) the pointwise error of the approximation from proposed hierarchical learning.
second row: the approximations at each level in the hierarchical learning. (left) the approximation $v(\cdot;\bm{\theta}_1^{\ast})$ at first level, (middle) the approximation $v(\cdot;\bm{\theta}_2^{\ast})$ at second level, (right) the target function for $v(\cdot;\bm{\theta}_2)$ at second level, which is equal to ($u_{\text{exact}}-v(\cdot;\bm{\theta}_1^{\ast})$).
}
\label{fig:advDiff2D_apprx_FourierEmbedNN}
\end{figure}

We compare the hierarchical learning method with the standard learning approach using the same size neural network with different Fourier feature embeddings; single embedding using $\sigma=2$ or $\sigma=5$, and multiple embeddings using $\sigma=2,5$. Fig.~\ref{fig:advDiff2D_losses_FourierEmbedNN} shows the training procedures in terms of three losses and relative $\mathcal{L}^2$-errors.  Among the standard learning experiments,  $\sigma=2$ embedding is suitable for this example as it has the most accurate approximation with a $\mathcal{L}^2$ error $4.76 \times 10^{-3}$. 
In comparison with the hierarchical learning method, hierarchical learning has the lowest error $2.41 \times 10^{-3}$ when the second level training is triggered after $1\times 10^4$ iterations of the first level. We also note that hierarchical learning converges after $4\times 10^4$ iterations, which is $2.5$ times faster than the other method. Fig.~\ref{fig:advDiff2D_apprx_FourierEmbedNN} shows the numerical solutions from both standard learning and hierarchical learning for reference.

\section{Discussions and conclusions}\label{sec:5.Discussion}

This study proposed a hierarchical learning method to solve PDEs using neural networks. We showed that a hierarchical design of networks is efficient and robust in representing a PDE solution that contains a wide range of scales. Various numerical tests have shown that a neural network has its characteristic scales, which have a fast convergence rate for training. The hierarchical approach we proposed in this study enables us to learn all possible scales in the solution. To impose a hierarchy, we tested two methods; 1) multi-layer perceptrons (MLPs) with various network complexities, and 2) Fourier feature embedded network. The first approach has a computational efficiency in solving a low complexity network while capturing the low-frequency components of the solution. The second
\clearpage

\noindent approach does not provide any computational gain as each network at different levels has the same complexity. Still, we can explicitly impose the range of scales of the solution through the Fourier feature embedded layer. The proposed hierarchical learning method has been tested through a suite of numerical tests, including linear and nonlinear PDEs, including the advection-diffusion problem with a multiscale velocity field. 

There are several issues to be addressed for the proposed hierarchical learning method. It is unclear to see the connection between the network complexity and its characteristic scales to represent a function. We have checked in our numerical experiments that changing the complexity of a network will change its corresponding scales. Still, we lack explicit and rigorous criteria to determine the characteristic scales. Also, we used the same network complexity for each level for the Fourier feature embedding approach, assuming that the Fourier embedded layer will determine its characteristic scales. We plan to investigate the effect of a network complexity that guarantees the imposed characteristic scales through the Fourier feature embedding. Another issue of our interest is the transition between different level networks. We have tested only a one-directional transition from large-scale to small-scale representation in the current study. In multigrid methods, several transition methods such as V or W cycles have shown successful accuracy results. We believe that various transition cycles will also improve the hierarchical learning method for solving PDEs, which we will report in another place. 

The numerical tests we considered here are elliptic PDEs, and we believe that the proposed method can apply to time-dependent problems. In solving a time-dependent problem, there are two approaches. One uses a network to learn the spatiotemporal scales at the same time. The other approach is to march the problem where the spatial variations are learned through a network \cite{raissi2019physics}. We are interested in designing hierarchical networks to resolve multiscale behaviors in the temporal domain, particularly to capture the long-time behavior of a dynamical system, such as the climatology of geophysical fluid systems. Lastly, we have tested the hierarchical learning method in the PINN framework in the current study. As the overarching idea of the proposed method is in the efficient representation of a multiscale function using hierarchical networks, we expect that the proposed method can apply to other network-based methods for solving PDEs, which we leave as future work.

\section*{Acknowledgments}
Yoonsang Lee is supported in part by NSF DMS-1912999 and ONR MURI N00014-20-1-2595.

\bibliographystyle{elsarticle-num}
\bibliography{references}

\end{document}